\documentclass[12pt]{iopart}
\usepackage[T1]{fontenc}
\usepackage[latin1]{inputenc}
\usepackage{float} 
\usepackage{graphicx}
\usepackage[margin=1in]{geometry}
\usepackage{setspace}
\usepackage{iopams}
\usepackage{amsmath,amssymb,amsfonts}
\usepackage{babel}
\usepackage{acronym}
\usepackage{units}
\usepackage{natbib}
\usepackage{hyperref}
\usepackage{fancyhdr}
\usepackage{wrapfig}
\usepackage{caption}
\usepackage{subfig}
\usepackage{multirow}
\usepackage{sidecap}
\usepackage{xcolor}

\renewcommand{\cite}{\citep}
\begin{document}

\title[Meta-learning \aclp{SNN} with Surrogate Gradient Descent]{Meta-learning \aclp{SNN} with Surrogate Gradient Descent}

\author{Kenneth M Stewart$^1$, Emre O Neftci$^2$} 
%\footnote{Present address:
% Department of Physics, University of Bristol, Tyndalls Park Road, 
% Bristol BS8 1TS, UK.} and Janet Williams$^3$}

\address{$^1$ Department of Computer Science, UC Irvine,
Irvine California, USA}
\address{$^2$ Department of Computer Science, Department of Cognitive Sciences,  UC Irvine \\ Peter Gr\"unberg Institute -- Neuromorphic Software Ecosystems\\ Forschungszentrum J\"ulich, Germany\\ Faculty of Electrical Engineering and Information Technology, RWTH Aachen, Germany\\}
\ead{kennetms@uci.edu}
%\eads{\mailto{#1}, \mailto{#2}}

% \author{Kenneth Stewart}
% \affil{Department of Computer Science, UC Irvine  }

% \author{Emre Neftci}
% \affil{Department of Computer Science, Department of Cognitive Sciences,  UC Irvine \\ Peter Gr\"unberg Institute -- Neuromorphic Software Ecosystems\\ Forschungszentrum J\"ulich, Germany\\ Faculty of Electrical Engineering and Information Technology, RWTH Aachen, Germany\\ }

\acrodef{IR}[IR]{Intrinsic Rewards and Motivation}
\acrodef{PPO}[PPO]{Proximal Policy Optimization}
\acrodef{RL}[RL]{Reinforcement Learning}
\acrodef{AC}[AC]{Arrenhius \& Current}
\acrodef{AD}[AD]{Automatic Differentiation}
\acrodef{AER}[AER]{Address Event Representation}
\acrodef{AEX}[AEX]{AER EXtension board}
\acrodef{AMDA}[AMDA]{``AER Motherboard with D/A converters''}
\acrodef{ANN}[ANN]{Artificial Neural Network}
\acrodef{API}[API]{Application Programming Interface}
\acrodef{BP}[BP]{Back-Propagation}
\acrodef{BPTT}[BPTT]{Back-Propagation-Through-Time}
\acrodef{BM}[BM]{Boltzmann Machine}
\acrodef{CAVIAR}[CAVIAR]{Convolution AER Vision Architecture for Real-Time}
\acrodef{CCN}[CCN]{Cooperative and Competitive Network}
\acrodef{CD}[CD]{Contrastive Divergence}
\acrodef{CG}[CG]{Computational Graph}
\acrodef{CMOS}[CMOS]{Complementary Metal--Oxide--Semiconductor}
\acrodef{CNN}[CNN]{Convolutional Neural Network}
\acrodef{COTS}[COTS]{Commercial Off-The-Shelf}
\acrodef{CPU}[CPU]{Central Processing Unit}
\acrodef{CV}[CV]{Coefficient of Variation}
\acrodef{CTC}[CTC]{connectionist temporal classification}
\acrodef{DAC}[DAC]{Digital--to--Analog}
\acrodef{DBN}[DBN]{Deep Belief Network}
\acrodef{DCLL}[DECOLLE]{Deep Continuous Local Learning}
\acrodef{DFA}[DFA]{Deterministic Finite Automaton}
\acrodef{DFA}[DFA]{Deterministic Finite Automaton}
\acrodef{divmod3}[DIVMOD3]{divisibility of a number by 3}
\acrodef{DPE}[DPE]{Dynamic Parameter Estimation}
\acrodef{DPI}[DPI]{Differential-Pair Integrator}
\acrodef{DSP}[DSP]{Digital Signal Processor}
\acrodef{DVS}[DVS]{Dynamic Vision Sensor}
\acrodef{EDVAC}[EDVAC]{Electronic Discrete Variable Automatic Computer}
\acrodef{EIF}[EI\&F]{Exponential Integrate \& Fire}
\acrodef{EIN}[EIN]{Excitatory--Inhibitory Network}
\acrodef{EPSC}[EPSC]{Excitatory Post-Synaptic Current}
\acrodef{EPSP}[EPSP]{Excitatory Post--Synaptic Potential}
\acrodef{eRBP}[eRBP]{Event-Driven Random Back-Propagation}
\acrodef{FPGA}[FPGA]{Field Programmable Gate Array}
\acrodef{FSM}[FSM]{Finite State Machine}
\acrodef{GPU}[GPU]{Graphical Processing Unit}
\acrodef{HAL}[HAL]{Hardware Abstraction Layer}
\acrodef{HH}[H\&H]{Hodgkin \& Huxley}
\acrodef{HMM}[HMM]{Hidden Markov Model}
\acrodef{HW}[HW]{Hardware}
\acrodef{hWTA}[hWTA]{Hard Winner--Take--All}
\acrodef{IF2DWTA}[IF2DWTA]{Integrate \& Fire 2--Dimensional WTA}
\acrodef{IF}[I\&F]{Integrate \& Fire}
\acrodef{IFSLWTA}[IFSLWTA]{Integrate \& Fire Stop Learning WTA}
\acrodef{INCF}[INCF]{International Neuroinformatics Coordinating Facility}
\acrodef{INRC}[INRC]{Intel Neuromorphic Research Community}
\acrodef{INI}[INI]{Institute of Neuroinformatics}
\acrodef{IO}[IO]{Input-Output}
\acrodef{IoT}[IoT]{internet of things}
\acrodef{IPSC}[IPSC]{Inhibitory Post-Synaptic Current}
\acrodef{ISI}[ISI]{Inter--Spike Interval}
\acrodef{JFLAP}[JFLAP]{Java - Formal Languages and Automata Package}
\acrodef{LIF}[LIF]{Linear Integrate and Fire}
\acrodef{LSM}[LSM]{Liquid State Machine}
\acrodef{LTD}[LTD]{Long-Term Depression}
\acrodef{LTI}[LTI]{Linear Time-Invariant}
\acrodef{LTP}[LTP]{Long-Term Potentiation}
\acrodef{LTU}[LTU]{Linear Threshold Unit}
\acrodef{LSTM}[LSTM]{Long Short-Term Memory}
\acrodef{MAML}[MAML]{Model Agnostic Meta Learning}
\acrodef{MCMC}{Markov Chain Monte Carlo}
\acrodef{MSE}{Mean-Squared Error}
\acrodef{NHML}[NHML]{Neuromorphic Hardware Mark-up Language}
\acrodef{NMDA}[NMDA]{NMDA}
\acrodef{NME}[NE]{Neuromorphic Engineering}
\acrodef{PCB}[PCB]{Printed Circuit Board}
\acrodef{PRC}[PRC]{Phase Response Curve}
\acrodef{PSC}[PSC]{Post-Synaptic Current}
\acrodef{PSP}[PSP]{Post--Synaptic Potential}
\acrodef{RI}[KL]{Kullback-Leibler}
\acrodef{RRAM}[RRAM]{Resistive Random-Access Memory}
\acrodef{RBM}[RBM]{Restricted Boltzmann Machine}
\acrodef{RTRL}[RTRL]{Real-Time Recurrent Learning}
\acrodef{ROC}[ROC]{Receiver Operator Characteristic}
\acrodef{RSA}[RSA]{Representational Similarity Analysis}
\acrodef{RDA}[RDA]{Representational Dissimilarity Analysis}
\acrodef{RDM}[RDA]{Representational Dissimilarity Matrix}
\acrodef{RNN}[RNN]{Recurrent Neural Network}
\acrodef{SAC}[SAC]{Selective Attention Chip}
\acrodef{SCD}[SCD]{Spike-Based Contrastive Divergence}
\acrodef{SCX}[SCX]{Silicon CorteX}
\acrodef{SG}[SG]{Surrogate Gradient}
\acrodef{SGD}[SGD]{Stochastic Gradient Descent}
\acrodef{SRM}[SRM]{Spike Response Model}
\acrodef{SNN}[SNN]{Spiking Neural Network}
\acrodef{STDP}[STDP]{Spike Time Dependent Plasticity}
\acrodef{SW}[SW]{Software}
\acrodef{sWTA}[SWTA]{Soft Winner--Take--All}
\acrodef{TPU}[TPU]{Tensorflow Processing Unit}
\acrodef{VHDL}[VHDL]{VHSIC Hardware Description Language}
\acrodef{VLSI}[VLSI]{Very  Large  Scale  Integration}
\acrodef{WTA}[WTA]{Winner--Take--All}
\acrodef{XML}[XML]{eXtensible Mark-up Language}

\begin{abstract}
Adaptive ``life-long'' learning at the edge and during online task performance is an aspirational goal of AI research.
Neuromorphic hardware implementing \acp{SNN} are particularly attractive in this regard, as their real-time, event-based, local computing paradigm makes them suitable for edge implementations and fast learning.
%Recent surrogate gradient descent learning rules that bridge machine learning methods with synaptic plasticity achieve competitive accuracy and performance on such hardware when compared to conventional \acp{ANN} of equivalent size. 
However, the long and iterative learning that characterizes state-of-the-art \ac{SNN} training is incompatible with the physical nature and real-time operation of neuromorphic hardware.
Bi-level learning, such as meta-learning is increasingly used in deep learning to overcome these limitations.
In this work, we demonstrate gradient-based meta-learning in \acp{SNN} using the surrogate gradient method that approximates the spiking threshold function for gradient estimations.
Because surrogate gradients can be made twice differentiable, well-established, and effective second-order gradient meta-learning methods such as \ac{MAML} can be used. 
We show that \acp{SNN} meta-trained using \ac{MAML} match or exceed the performance of conventional \acp{ANN} meta-trained with \ac{MAML} on event-based meta-datasets. 
Furthermore, we demonstrate the specific advantages that accrue from meta-learning: fast learning without the requirement of high precision weights or gradients. 
Our results emphasize how meta-learning techniques can become instrumental for deploying neuromorphic learning technologies on real-world problems.
\end{abstract}

\maketitle


\section{Introduction}
Rapid adaptation to unfamiliar and ambiguous tasks are hallmarks of cognitive function and long-standing goals of Artificial Intelligence (AI). 
Neuromorphic electronic systems inspired by the brain's dynamics and architecture strive to capture its key properties to enable low-power, versatile, and fast information processing \cite{Mead90_neurelec,Indiveri_etal11_neursili,Davies19_bencprog}. 
Several recent neuromorphic systems are now equipped with on-chip local synaptic plasticity dynamics \cite{Chicca_etal13_neurelec,Pfeil_etal12_4-bisyna,Davies_etal18_loihneur}.
Such neuromorphic learning machines hold promise to building fast and power-efficient life-learning machines \cite{Neftci18_datapowe}.

\acp{SNN} can be modeled as a special case of artificial \acp{RNN} with internal states akin to the \ac{LSTM} \cite{Neftci_etal19_surrgrad}. 
Using a surrogate gradient approach that approximates the spiking threshold function for gradient estimations, \acp{SNN} can be trained to match or exceed the accuracy of conventional neural networks on event-based vision, audio, and reinforcement learning tasks \cite{Kaiser_etal19_synaplas,Cramer_etal20_traispik,Bellec_etal19_biolinsp,Bohnstingl_etal20_onlispat,Zenke_Ganguli17_supesupe,Neftci_etal17_evenranda}. 
Although these methods achieve state-of-the-art accuracy in \acp{SNN}, they are not practically realizable on neuromorphic hardware or any other online learning systems, for several reasons:
%Correlation
Firstly, learning via (stochastic) gradient descent requires data to be sampled in an independent and identically distributed fashion \cite{Vapnik13_natustat}. 
However, when sensory data is acquired and processed during task performance, data samples are generally correlated, leading to many convergence problems, including catastrophic forgetting~\cite{McClelland_etal95_whyther}. 
%Learning rates
Secondly, many networks use batch sizes larger than one. 
While networks with a batch size equal to one eventually converge \cite{LeCun_Bottou04_largscal}, using a smaller batch size means that learning rates must be smaller as well. Smaller learning rates result in smaller weight updates which require memories or buffers that store the weights or weight updates with higher precision, and hence hardware area.
%Data inefficiency
Thirdly, surrogate gradient-based \ac{SNN} training inherits other fundamental issues of deep learning, namely that very large datasets and a large number of iterations are necessary for convergence.
The combination of the three problems stated above, \emph{i.e.}  correlated data samples, data inefficiency, and memory requirements hamper the successful deployment of neuromorphic hardware to solve real-world learning problems.

\begin{SCfigure}
  \includegraphics[width=0.45\textwidth]{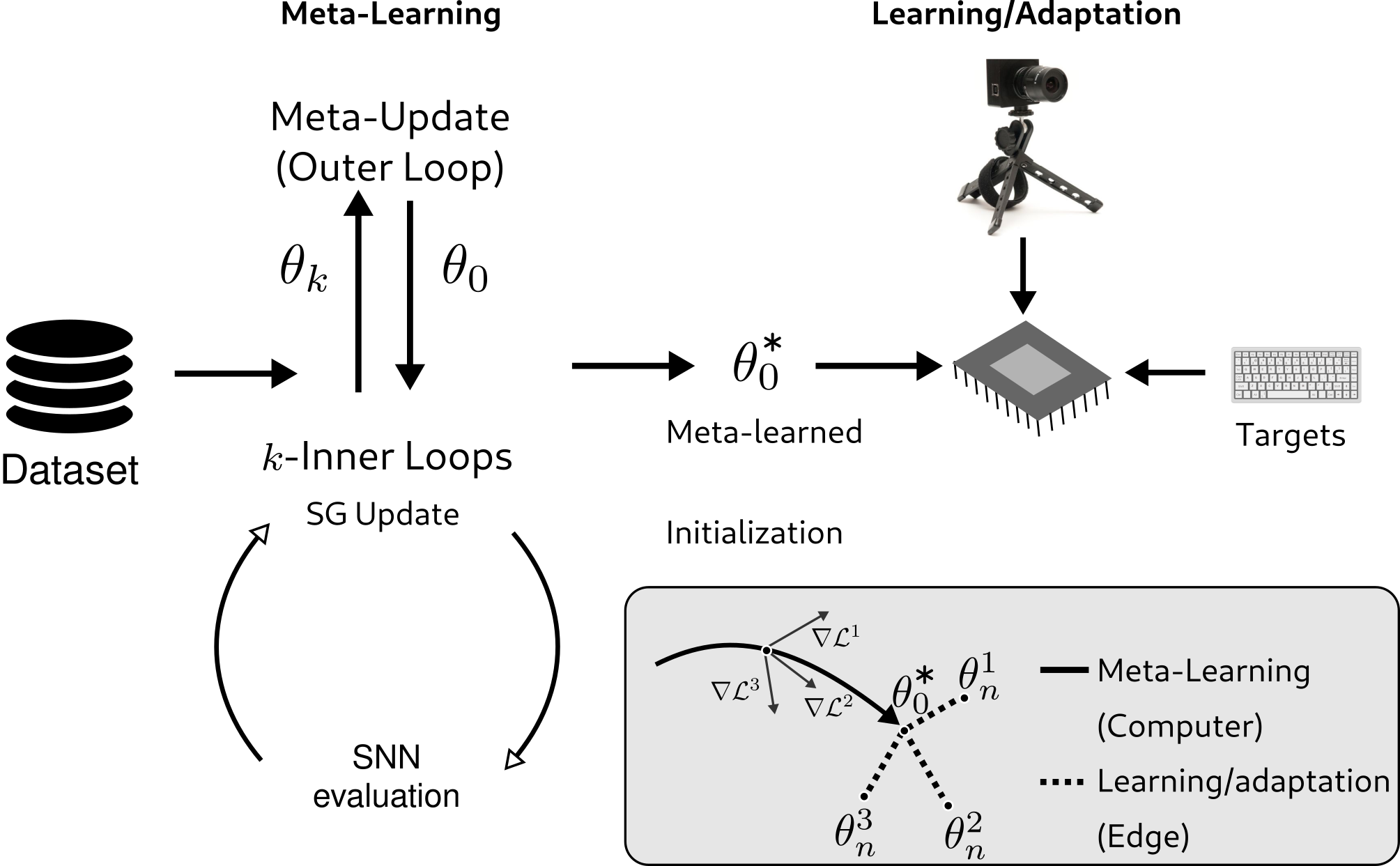}
  \caption{\label{fig:introduction_figure} Meta-Learning for SNNs using Surrogate Gradients.
  In the first phase, an SNN or a functional simulator of a neuromorphic hardware's \acp{SNN} is meta-trained using surrogate gradient methods on a class of tasks $T_i$ stored on a computer. The goal of meta-training is to learn an initial parameter set $\theta_0^\ast$ such that out-of-sample tasks (\emph{e.g.} $\theta_n^1$, $\theta_n^2$, $\theta_n^3$, ...$\theta_n^\ast$) can be learned quickly. In the envisioned application, $\theta_0^\ast$ would be learned offline on a conventional computer and learning/adaption would take place at the edge, using neuromorphic sensing and processing.}
\end{SCfigure}

%Expand
% Animals arguably learn in several stages. 
% \textbf{describe here learning along the lines of Lecun's arguments here \url{http://helper.ipam.ucla.edu/publications/mlpws4/mlpws4_15927.pdf}}.
% These multiple stages of learning motivate the idea of pre-training networks, such that learning of new concepts occur only after sufficient "background" knowledge has been acquired \cite{Lake_etal17_builmach}.
In this article, we demonstrate that gradient-based meta-learning on \acp{SNN} can solve these problems in practical cases with technological interest, and are particularly well suited to the constraints of neuromorphic hardware and online learning (Fig. \ref{fig:introduction_figure}).
To do so we combine \ac{MAML}, a second-order gradient-based method that optimizes the network hyperparameters, and the surrogate gradient method \cite{Neftci_etal19_surrgrad}.
Two ingredients were key to the results of our work.
First, surrogate functions used to estimate \ac{SNN} gradients can be made twice differentiable, hence are suitable for second-order learning as in \ac{MAML}.
Second, the definition of suitable event-based datasets to demonstrate meta-learning on \acp{SNN}.
While \ac{MAML} had been previously applied to \acp{SNN}, prior work focused on meta-training Hebbian \ac{STDP} dynamics on non-event-based datasets which do not take any advantage of the event-based nature of \acp{SNN}.  
Furthermore, surrogate gradient learning implementing stochastic gradient descent can be implemented as a form of three-factor learning \cite{Gerstner_etal18_eligtrac,Zenke_Ganguli17_supesupe,Kaiser_etal20_synaplas,Bellec_etal19_biolinsp} that vastly exceeds the performance of classical \ac{STDP}, while being compatible with neuromorphic hardware implementations \cite{Payvand_etal20_errothre,Cramer_etal20_traispik}. 
The meta-training of \acp{SNN} using the surrogate gradient method can be seen as a tool to adapt and tune synaptic plasticity circuits.

We study the \ac{SNN} \ac{MAML} approach in the context of few-shot learning, whereby a model is trained on a set of labeled tasks drawn from a given \emph{domain} of tasks to adapt to unseen ones of the same domain using a small number of samples and iterations. 
Examples of few-shot learning are learning novel hand or body gestures, agents learning to take new goal-driven actions in a new maze, or optimizing automatic speech recognition to the individual pronunciation of the subject. 

One important obstacle to meta-learning research in neuromorphic engineering is the lack of suitable datasets. 
Neuromorphic hardware implementing \acp{SNN} is most suitable to processing event-based datasets and loses most of its salient features when applied to static data \cite{Davies19_bencprog}. 
The Omniglot \cite{Lake_etal17_builmach} and MiniImagenet \cite{Vinyals_etal16_matcnetw} datasets have been pivotal in pushing the field of meta-learning ahead.
However, there exists no event-based dataset that is comparable to Omniglot or MiniImagenet where modeling dynamics are crucial to solving the problem.
Taking inspiration from existing meta-learning benchmarks that fuse multiple datasets \cite{mulitdigitmnist}, we define new benchmarks that consist of combinations of event-based datasets recorded using neuromorphic vision sensors. 
We demonstrate performances that match - and in some cases exceed - the performance of conventional neural networks trained on these datasets.

Finally, we analyze the updated statistics, which reveal that the \ac{MAML} not only results in fast learning but does so using large weight updates. These results lay promise for online learning using low-precision weight memory.

\subsection*{Specific Contributions}
This work provides 1) a method of parameter initialization that enables neuromorphic hardware to few-shot learn new tasks; 2) a method to construct meta-datasets using data taken from the DVS neuromorphic sensor, with two examples made publicly available: Double NMNIST and Double ASL-DVS; and 3) the effectiveness of second-order meta-training of \acp{SNN}. 
We present this work as a stepping stone towards implementing MAML with SNNs in neuromorphic hardware for fast adaptation to streamed event-based sensor data.

\section{Methods}
\subsection{Model Agnostic Meta-Learning}

Define a neural network model $y = f(x,\theta)$ that produces an output batch $y$ given its parameters $\theta$ and an input batch $x$.
For simplicity, we focus here on classification problems, such that $y$ represents logits and $\arg\max y$ is a class, although any supervised learning problem would be suitable.
In the classification case, each batch consists of $K$ samples of each class.
% In a conventional image recognition network, $x$ is a 3rd order tensor (omitting batch dimension, one channel per color).
% In this work, an SNN takes raw neuromorphic sensor data as input.
% In this case, the input is typically a 4th order tensor (time, polarity, height, width), where polarity indicates the increase or decrease of intensity.
%In both cases, 
The parameters $\theta$ are trained by minimizing a task-relevant loss function $\mathcal{L}(f(x,\theta),t)$, such as cross-entropy, where $t$ is a batch of targets. 
%$f$ can be trained 

The goal of meta-learning is to optimize the meta-parameters of $f$, such as the initialization parameters noted as $\theta_0$.
This work makes use of the standard second-order \ac{MAML} algorithm to meta-train the \ac{SNN}.
The standard MAML workflow is designed to optimize the parameters of a neural network model across multiple tasks in a few-shot setting. 
MAML achieves this using two nested optimization loops, one ``inner'' loop and one ``outer'' loop. 
The inner loop consists of a standard \ac{SGD} update, where the gradient operations are traced for auto-differentiation \cite{Griewank_Walther08_evalderi}. 
In the outer loop, an update is made using gradient descent on the meta-parameters.

To make use of \ac{MAML}, it is essential to set up the experimental framework accordingly.
Define three sets of tasks:  meta training $\mathcal{T}^{trn} $, meta validation $\mathcal{T}^{val} $, and meta testing $\mathcal{T}^{tst} $. 
Each task consists of a training dataset $\mathcal{D}^{trn}$ and a validation dataset $\mathcal{D}^{val}$ of the form $\mathcal{D}=\{x_i,t_i\}_{i=1}^M$.
Here $x_i$ denotes the input data, $t_i$ the target (label) and M is the number of target samples.
In general, the datasets corresponding to different tasks can have different sizes, but we omit this in the notation to avoid clutter.
%where $x_i$ is data and $t_i$ is the target (label) and $M$ is the number of data and target samples 
%(although datasets may have different sizes, we omit this in the notation to avoid clutter). 
During learning, a task is sampled from $\mathcal{T}^{trn}$ and $N$ inner loop updates are made using batches of data sampled from $\mathcal{D}^{trn}$. 
The resulting parameters $\theta_{N}$ are then used to make the outer loop update using the matching validation dataset $\mathcal{D}^{val}$.
During each inner loop update, one or more \ac{SGD} update steps are performed over a task-relevant loss function $\mathcal{L}$:
\begin{equation}\label{eq:inner_loop}
\begin{split}
    \theta_{n+1}(\theta_n, \mathcal{D}^{trn}) &= \theta_{n} - \alpha \nabla_{\theta_{n}} \mathcal{L}(f(x,{\theta_{n}}), t), \\
    &\text{where }\{x,t\}\in\mathcal{D}^{trn}\text{ for }n=1,...,N.
\end{split}
\end{equation} 
Here $n$ is the number of inner loop adaptation steps, $\alpha$ is the inner loop learning rate. 
Note the dependence of $\theta_{n+1}$ on the initial parameter set $\theta_0$ at the beginning of the inner loop through the recursion.
$\nabla_{\theta_{n}}$ here indicates the gradient over the inner loop loss $\mathcal{L}$ on the network using parameters $\theta_{n}$.
The outer loop loss is defined as:
\begin{equation}
    \mathcal{L}_{outer}(\theta_0) = \sum_{T \in P(\mathcal{T}^{trn})} \mathcal{L}(f(\theta_N(\theta_0, \mathcal{D}^{trn}), \mathcal{D}^{val})),
\end{equation}
where $T=\{\mathcal{D}^{trn}, \mathcal{D}^{val}\}$. Note that in practice the above expression is generally computed over a random subset of tasks rather than the full set $\mathcal{T}^{trn}$.
Notice that the outer loop loss is computed over the validation dataset $\mathcal{D}^{val}$, whereas $\theta_N(\theta_0)$ is computed using the training dataset  $\mathcal{D}^{trn}$, which is argued to improve generalization.
The goal is to find the optimal $\theta_0$, denoted $\theta_0^\ast$ such that:
\begin{equation}
    \theta_0^\ast = \arg\min_{\theta_0} \mathcal{L}_{outer}(\theta_0).
\end{equation}
Provided the inner loop loss is at least twice differentiable with respect to $\theta_0$, the optimization can be performed via gradient descent over the initial parameters $\theta_0$, using a standard gradient-based optimizer using gradients
$\nabla_{\theta_0}\mathcal{L}_{outer}(\theta_0).$
Successive applications of the chain rule in the expression above results in second order gradients of the form $\nabla_{\theta_n}^2 \mathcal{L}(f(x,\theta_{n}),t)$. 
If these second-order terms are ignored, it is still possible to meta-learn \cite{Finn_etal17_modemeta}, using the method called first-order MAML (FOMAML).

In our experiments, we use the ADAM optimizer for the outer loop loss function and vanilla \ac{SGD} for the inner loop loss.
This choice is motivated by a hybrid learning framework whereby the outer loop training can occur offline with large memory and compute resources (e.g. ADAM which requires more memory and compute), whereas the inner loop is constrained by hardware at the edge.
The model is validated and tested on $\mathcal{D}^{val}$ and $\mathcal{D}^{tst}$, respectively.
In the following, we describe the \ac{SNN} model used with \ac{MAML}.

\subsection{\ac{MAML}-compatible Spiking Neuron Model}
The neuron model used in the \acp{SNN} in our work follows Leaky Integrate \& Fire (LIF) dynamics as described in \cite{Kaiser_etal20_synaplas}. 
For completeness, we summarize the dynamics of the neuron model here:
\begin{equation}\label{eq:lif_equations}
  \begin{split}
    u_i^{t} &= \sum_j w_{ij} p_j^{t} - \rho r_i^{t}  + b_i, \\
    s_i^{t} &= \Theta( u_i^{t}), \\
    p_j^{t+\Delta t} &= \alpha p_{j}^{t} + (1-\alpha) q_{j}^{t}, \\
    q_j^{t+\Delta t} &= \beta  q_{j}^{t} + (1-\beta ) s_{j}^{t}, \\
    r_i^{t+\Delta t} &= \gamma r_{i}^{t} + (1-\gamma) s_{i}^{t}, \\
  \end{split}
\end{equation}
where $u_i$ is the membrane potential, $w_{ij}$ are the synaptic weights between pre-synaptic neuron $j$ and post-synaptic neuron $i$ and $\Delta t$ is the timestep.
Neurons emit a spike $s_i^{t}$ at time $t$ when the threshold of their membrane potential $\Theta(u_i^{t})$ is reached.
$\Theta(u_i^{t})$ is the unit step function, where $\Theta(u_i) = 0$ if $u_i < u_{th}$, otherwise $1$.
$p$ and $q$ describe the traces of the membrane potential of the neuron and the current of the synapse, respectively.
For each incoming spike to a neuron, each trace undergoes a jump of height 1 and decays exponentially 
%with a time constant
% of $\tau_{mem}$ for $P$ and $\tau_{syn}$ for $Q$ 
if no spikes are received.
The constants
\begin{equation}
\alpha=\exp(-\frac{\Delta t}{\tau_{\mathrm{mem}}}), \beta =\exp(-\frac{\Delta t}{\tau_{\mathrm{syn}}})\text{ and }\gamma =\exp(-\frac{\Delta t}{\tau_{\mathrm{rfr}}})
\end{equation}
reflect the time constants of the membrane $p$, synaptic $q$, and refractory $r$ dynamics. %membrane potential $U$.
Weighting the trace $p_j$ with the synaptic weight $w_{ij}$ results in the Post-Synaptic Potential (PSP) of post-synaptic neuron $i$ caused by pre-synaptic neuron $j$.
The constant $b_i$ is a bias current representing the intrinsic excitability of the neuron.
The reset mechanism is captured by the dynamics of $r_i$, and the factors $\tau_{mem}$,$\tau_{syn}$ and $\tau_{ref}$ are time constants of the membrane, synapse, and reset dynamics respectively. 
Note that Eq.\ref{eq:lif_equations} is equivalent to a discrete-time version of the \ac{SRM} with linear filters \cite{Gerstner_Kistler02_spikneur}.

To compute the second-order gradients, \ac{MAML} requires the SNN to be twice differentiable. 
However, the spiking function $\Theta$ is non-differentiable. 
The surrogate gradient approach, where $\Theta$ is replaced by a differentiable surrogate function $\sigma$ for computing gradients, has been used to successfully side-step this problem \cite{Neftci_etal19_surrgrad}.
For \ac{MAML}, the surrogate function can be chosen to be a twice differentiable function. 
Although many suitable surrogate gradient functions exist, the fast sigmoid function described in \cite{Zenke_Vogels20_remarobu} strikes a good trade-off between simplicity and effectiveness in learning and is twice differentiable:
$\sigma'(x) := \frac{1}{(\beta |x|+1)^2}$.
All simulations in this work use the fast sigmoid function as a surrogate function.

Because \acp{SNN} is a special case of recurrent neural networks, it is possible to apply Automatic Differentiation tools for implementing the gradient \cite{Baydin_etal17_autodiff,Paszke_etal17_autodiff}.
This also applies to the calculation of the second-order gradients needed for backpropagating the gradient of the inner loss. 

%
% \begin{equation}\label{eq:clif}
%   \begin{split}
%     U_i(t) =& V_i(t) - \rho R_i(t) + b_i,\\
%     \tau_{mem}\frac{\mathrm{d}}{\mathrm{d}t} V_i(t) =& - V_i(t) + I_i(t),\\
%     \tau_{ref} \frac{\mathrm{d}}{\mathrm{d}t} R_i(t) =& -R_i(t)  +  S_i(t),\\
%   \end{split}
% \end{equation}
%
% where $S_i(t)=\sum_f \delta (t-t_i^f)$ represents the spike train of neuron $i$ spiking at times $t^f_i$ and $\delta$ is the Dirac delta function.

% $I_i$ denotes the total synaptic current expressed as:
% \begin{equation}\label{eq:current}
%     \begin{split}
%       \tau_{syn} \frac{\mathrm{d}}{\mathrm{d} t} I_{i}(t)  = & -I_{i}(t) + \sum_{j\in \text{pre}} W_{ij} S_j(t),
%     \end{split}
% \end{equation}

% where $W_{ij}$ are the synaptic weights between pre-synaptic neuron $j$ and post-synaptic neuron $i$.
% Because the membrane potential $V_i$ and current $I_i$ are linear with respect to the weights $W_{ij}$, the dynamics of $V_i$ can be rewritten as:
% \begin{equation}\label{eq:linear}
%     \begin{split}
%       V_i(t) =& \sum_{j\in \text{pre}} W_{ij} P_{j}(t), \\
%       \tau_{mem} \frac{\mathrm{d}}{\mathrm{d} t} P_{j}(t)  = & -P_{j}(t) +  Q_{j}(t),\\
%       \tau_{syn} \frac{\mathrm{d}}{\mathrm{d} t} Q_{j}(t)  = & -Q_{j}(t) +  S_{j}(t).
%     \end{split}
% \end{equation}
% where 

% The dynamics described in Eq. \ref{eq:clif} and Eq. \ref{eq:linear} can be expressed in discrete time as:

\subsection{Datasets}

 \begin{SCfigure}
    \includegraphics[width=.5\linewidth]{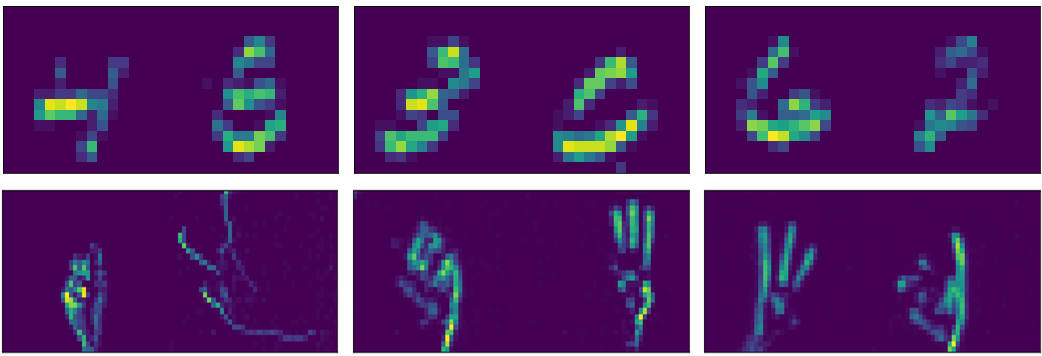}
    \centering
    \caption{Top: Examples of Double N-MNIST tasks. Each sample contains a combination of two N-MNIST digits to make two-digit numbers. Bottom: Examples of Double ASL-DVS tasks. Each sample contains a combination of two ASL-DVS letters. In all examples, the images are DVS events summed over 100ms into frames.}
    \label{fig:double_nmnist}
\end{SCfigure}

We benchmark our models on modifications of datasets collected using event-based vision sensors \cite{Lichtsteiner_etal08_128x120,Posch_etal11_qvga143}, the Neuromorphic MNIST (N-MNIST) \cite{Orchard_etal15_convstat}, and the American Sign Language Dynamic Vision Sensor (ASL-DVS) \cite{bi2019graph} datasets.
NMNIST consists of $32\times 32$, $300$ms long event data streams of MNIST images recorded with an ATIS Camera \cite{Posch_etal11_qvga143}. 
The dataset contains 60,000 training event streams and 10,000 test event streams.
From the N-MNIST dataset, we create Double N-MNIST datasets.
Each event stream of Double N-MNIST is a combination of two N-MNIST event streams to make $64\times 32$, $300$ms long event data streams of double-digit numbers that are downsampled to $32\times 16$, $100$ms long event data streams.
Because there are ten digits in the original N-MNIST dataset, 100 different double-digit numbers can be created.
These 100 different numbers can be used to create a meta-dataset with K=100 tasks, where each double-digit number represents one task.
%Because there are ten digits, 100 different tasks where each task is a double digit number can be used to create N-shot K-way datasets with each double digit representing a task.
We create N-shot K-way meta training, meta validation, and meta-test Double N-MNIST datasets from the training and test N-MNIST dataset.
Each meta dataset consists of a subset of the 100 total possible tasks.
The meta training dataset contains 64 tasks, the meta validation dataset contains 16 tasks, and the meta test dataset contains 20 tasks.

The ASL-DVS dataset contains 24 classes corresponding to letters A-Y, excluding J, in American Sign Language recorded using a DAVIS 240C event-based sensor \cite{Brandli_etal14_240180}. 
Data recording was performed in an office environment under constant illumination. 
The dataset contains 4200 $240\times 180$ $100$ms long event data streams of each letter, for a total of 100,800 samples.
Like Double N-MNIST, each event stream of Double ASL-DVS is a combination of two ASL-DVS event data streams to make $480\times 360$, $100$ms long event data streams of two letters that are downsampled to $80\times 30$, $100$ms long event data streams.
Out of the 24 classes, 576 different tasks consisting of double ASL letter event streams can be used to create N-shot K-way meta datasets with each double ASL letter representing a task.
From the ASL-DVS dataset, we create Double ASL-DVS N-shot K-way meta training, validation, and test datasets.
The meta training dataset contains 369 tasks, the meta validation dataset contains 92 tasks, and the meta test dataset contains 115 tasks.
Example images from the Double N-MNIST and Double ASL-DVS datasets are shown in Fig. \ref{fig:double_nmnist}.

For all datasets, gradients were computed for the last $30\mathrm{ms}$ of the sequence to reduce the memory footprint. 

\subsection{Model Architecture}
\begin{table}[!h]
%\small
\begin{center}
\caption{SNN MAML Network Architecture \label{tab:arch}}
\def\x{$\times$}
\scalebox{1}{
\begin{tabular}{|c|c|c|c|} \hline
Layer & Kernel & NMNIST Output    & ASL-DVS Output\\ \hline
input &      & 32\x16\x2 & 80\x30\x2\\ 
1     & 32c5p0s1   & 32\x16\x32  &80\x30\x32\\
3     & 2a     & 16\x8\x32  &40\x15\x32\\
4     & 64c5p0s1   & 16\x8\x64  &40\x15\x64\\
5     & 2a     & 8\x4\x64    &20\x7\x64\\
8     & 128c5p0s1   & 8\x8\x128  &20\x7\x128\\
9     & 2a   & 4\x2\x128  &10\x3\x128\\
output     & -      & K$=$5       &K$=$5\\  \hline
\end{tabular}}
\end{center}
\scriptsize{Notation: \verb~Ya~ represents \verb~YxY~ max pooling, \verb~XcYpZsS~ represents \verb~X~ convolution filters (\verb~YxY~) with padding $Z$ and stride $S$. }
\normalsize
\vspace{-3mm}
\end{table}

The architecture for the models trained is shown in Table \ref{tab:arch}. 
The architecture for all models trained is equivalent, consisting of three convolutional layers and a linear output layer.
SNN output membrane potentials are encoded into classes by using the output neuron with the highest membrane potential as the classification.

\section{Results}
\subsection{Few-shot Learning Performance on Double NMNIST and Double ASL Tasks}
% First, for each dataset, we validate that the parameters of the meta trained model are indeed usable by a conventional (non meta-trained) SUGR model for few-shot adaptation to new data. 
% The advantage of transferring the initial parameter set $\theta_o^*$ are 1) updating the model with the transferred features only requires one backwards pass instead of two which is computationally less expensive; 2) the meta-trained $\theta_o^*$ can be used to learn an out of sample task in only a few shots; and 3) when implemented in neuromorphic hardware, the module-task dependence parameters $\sigma$ can prioritize plasticity in the Loihi neural compartments.
For each dataset, we ran 1-shot 5-way learning experiments on models trained using MAML.
The MAML models were meta-trained on the meta training tasks ($\mathcal{T}^{trn}$) with the meta validation tasks ($\mathcal{T}^{val}$) used to compute the loss gradient in the outer loop. 
The meta-trained models were then tested on the meta-test tasks ($\mathcal{T}^{tst}$).
% Each model consists of three convolution layers with a linear output layer.
A summary of our results for each dataset is shown in Table \ref{tab:results}.
The results in Table \ref{tab:results} are obtained from averaging the inference performance of a meta-trained model over 10 trials on the test datasets ($D^{tst}$) of the meta validation and meta test tasks.
Each trial has different random batches of data sampled from training tasks $\mathcal{T}^{trn} $ and validation tasks $\mathcal{T}^{val} $ respectively.
All experiments only used a single inner loop gradient step (\emph{i.e.} in \ac{MAML} $N$ was set to $1$).
Additionally, we compare the results of the SNNs to equivalent non-spiking models meta-trained with MAML as well as SNNs trained with first-order MAML. 
First-order MAML ignores all terms involving the second-order gradients and is thus similar to the joint training of the tasks. 
This has the advantage of reducing the memory footprint required for learning but is known to reduce the accuracy of the meta-trained model \cite{Finn_etal17_modemeta}.
For the non-spiking models, the input data is first converted from the address event representation to static images by summing the events over the time dimension.
%When inputting the datasets into the non-spiking MAML models,
%the address events in the event stream are summed across the time dimension to create images.

The results show that both spiking and non-spiking MAML achieve 1-shot learning performance on the datasets comparable to the state-of-the-art performance shown by non-meta models.
The state-of-the-art test accuracy for standard non-meta model training on the NMNIST dataset with SNNs is $99.2\pm 0.02$\% accuracy \cite{Shrestha_Orchard18_slayspik}.
The SNN achieves $98.23\pm 1.12$\% test accuracy on Double NMNIST in a one-shot learning scenario.
%The state of the art test accuracy for standard training on the ASL-DVS dataset is 90.1\% test accuracy \cite{bi2019graph}. 
Our SNN network achieves a test accuracy on the Double ASL-DVS dataset of $96.04\pm 2.31$\%.
For both datasets, first-order \ac{MAML} performed significantly worse, highlighting the importance of differentiable surrogate gradients for successful meta-training.

On average MAML on SNNs tend to match or outperform non-spiking MAML on event-based datasets.
This is likely because the dynamics of the SNN neurons are well suited for processing and learning the Spatio-temporal patterns of the event-based data streams the datasets are composed of.

% \usepackage{multirow}
\begin{table}[!ht]
\begin{center}
\caption{1-Shot 5-Way Accuracy Results. Train accuracy indicated accuracy over the test datasets ($D^{tst}$) in the meta validation set $\mathcal{T}^{trn}$ and Test Accuracy indicates accuracy on the meta test set $\mathcal{T}^{trn}$.
\label{tab:results}}
%\resizebox{\textwidth}{!}{%
\begin{tabular}{|l|r|l|l|}
\hline
Task                   & \multicolumn{1}{l|}{Algorithm} & Train Accuracy  & Test Accuracy \\ \hline\hline\hline
\multirow{3}{*}{Double N-MNIST} & MAML (SNN)              & 98.76$\pm$1.05\%     & 98.23$\pm$1.12\%            \\ \cline{2-4} 
                                & MAML (CNN)              & 99.09$\pm$.53\%     & 98.35$\pm$1.26\%          \\ \cline{2-4}
                                & FOMAML (SNN)            & 92.59$\pm$1.0\%     & 92.63$\pm$.74\%          \\ \hline\hline
\multirow{3}{*}{Double ASL DVS} & MAML (SNN)              & 95.77$\pm$.88\%     & 96.04$\pm$2.31\%          \\ \cline{2-4}
                                & MAML (CNN)              & 94.93$\pm$.92\%     & 94.97$\pm$1.63\%          \\ \cline{2-4}
                                & FOMAML (SNN)            & 94.97$\pm$1.12\%     & 94.27$\pm$.6\%          \\ \hline
\end{tabular}%
%}
%% MAML Double N-MNIST Test:
%run test.py logs/runs2/snn_notime/2021-05-18_141456/config.json --use-cuda

\end{center}
\end{table}

\subsection{Generalization of Learning Performance}
\begin{figure}
\centering
    \includegraphics[width=.45\linewidth]{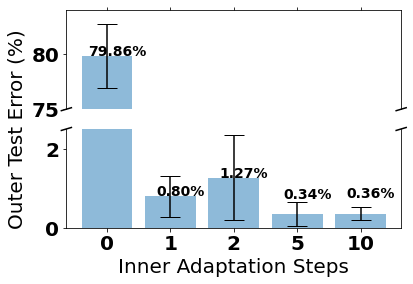}
    \includegraphics[width=.45\linewidth]{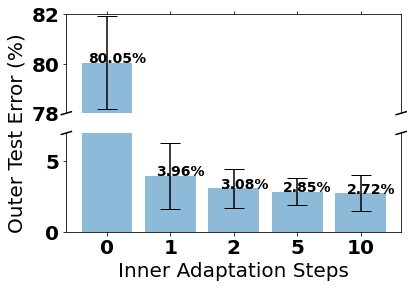}
\caption{Example of how changing the number of inner loop training steps during meta-testing affects the error of the meta-trained model. Accuracy increases as the number of gradient steps increases and without adapting to a new task the model will have very high error. Left: Double NMNIST. Right: Double ASL-DVS.}
\label{fig:adaptation_steps}
\end{figure}

\begin{figure}
\centering
    \includegraphics[width=.45\linewidth]{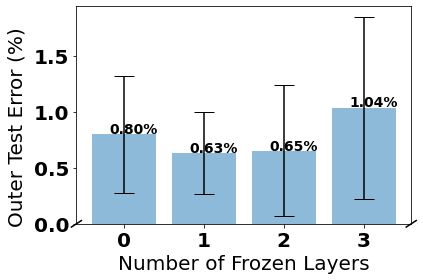}
    \includegraphics[width=.45\linewidth]{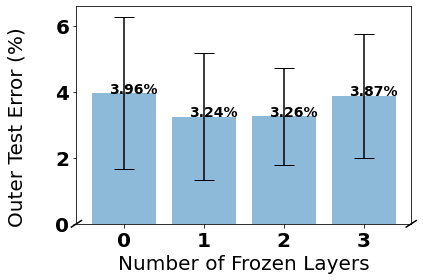}
\caption{Example of how freezing layers of the network during inner loop adaptation does not greatly impact learning performance. This supports the network is using feature reuse as in \cite{Raghu2020Rapid}. Left: Double NMNIST. Right: Double ASL-DVS.}
\label{fig:freeze_layers}
\end{figure}

MAML requires selecting hyper-parameters such as the number of update steps and the learning rates.
In real-world scenarios, the input constraints cannot be tightly controlled, leading to potential mismatches with the MAML hyper-parameters.
For example, in a real-time gesture learning scenario, the parameter update schedule may not be tightly linked to the time the gesture is presented.
Here we study the ability of MAML trained \acp{SNN} to generalize across different input conditions.

The ability of MAML to generalize learning performance across its settings, such as the number of steps, has been previously documented for conventional \acp{ANN} \cite{finn2018metalearning}.
 %\textbf{Example: in a gesture learning scenario, the parameter update schedule may not be tightly linked to the gesture presentation.}
 
Here, we demonstrate that this feature extends to our \ac{SNN}.
Using an SNN MAML network meta-trained on the Double NMNIST dataset, we varied the number of gradient steps during inner loop adaptation on test data.
The Fig. \ref{fig:adaptation_steps} shows how changing the number of gradient steps during inner loop adaptation affects the one-shot 5-way learning performance on each dataset.
On both datasets, as the number of inner loop gradient steps increases the performance increases.
Therefore there is a trade-off between the computational overhead of performing multiple gradient steps during one-shot learning and accuracy.

We also show how the learning performance is affected when layers of the network are frozen during few-shot learning.
Using a network meta-trained on the Double NMNIST dataset, we progressively froze layers of the network to observe the impact on performance, which is shown in Fig. \ref{fig:freeze_layers}.
Even when all layers of the network were frozen, in this case, three, there is not a significant impact on performance.
This gives further evidence to the claim that MAML learns a suitable representation for few-shot learning instead of rapid learning \cite{Raghu2020Rapid}.
This is interesting from an engineering perspective, as the network with a meta-learned initialization can achieve high performance on learning new tasks with only one gradient update and only at the final layer.
This is well suited for real-time adaptation in neuromorphic hardware demonstrated in previous work \cite{Stewart_etal20_onlifew-}.

\subsection{MAML Few-shot Learning Relies on Few, Large Magnitude Updates} \label{sec:update_magnitude}
To obtain adequate generalization, conventional deep learning relies on many, small magnitude updates across a large dataset. 
This is achieved using a relatively small learning rate. 
The usage of small learning rates is challenging on a physical substrate, as it requires high precision memory to accumulate the gradients across updates.
This problem is further compounded by the fact that learning on a physical substrate cannot be easily performed using batches of samples.

Interestingly, few-shot learning has the opposite requirements: few but large magnitude updates.
This result is extremely relevant for neuromorphic hardware which uses low precision parameters.
%In emerging resistive switching devices, such as memristors, both dynamic range and the resolution of the conductances are limited \cite{}. 
%
\begin{table}[!t]
\caption{Comparison of the Magnitude of Updates Between MAML and non-MAML Learning}
\label{tab:mag}
\resizebox{\textwidth}{!}{%
\begin{tabular}{|l|l|l|l|l|}
\hline
Algorithm            & Avg Magnitude & Sum of Magnitudes  & Max Magnitude \\ \hline
MAML Outer Loop & 7.72e-05$\pm$.0002   & 0.296$\pm$0.599     &  .002$\pm$.0017  \\ \hline
MAML Inner Loop & .0044$\pm$.0007      & 17.03$\pm$2.69      & 0.1548$\pm$0.16     \\ \hline
Non-Meta              & 0.0005$\pm$.0002    & 0.5499$\pm$0.2412       &  0.0011$\pm$.0002    \\ \hline
\end{tabular}%
}
\end{table}
\begin{figure}
\centering
    \includegraphics[width=.45\linewidth]{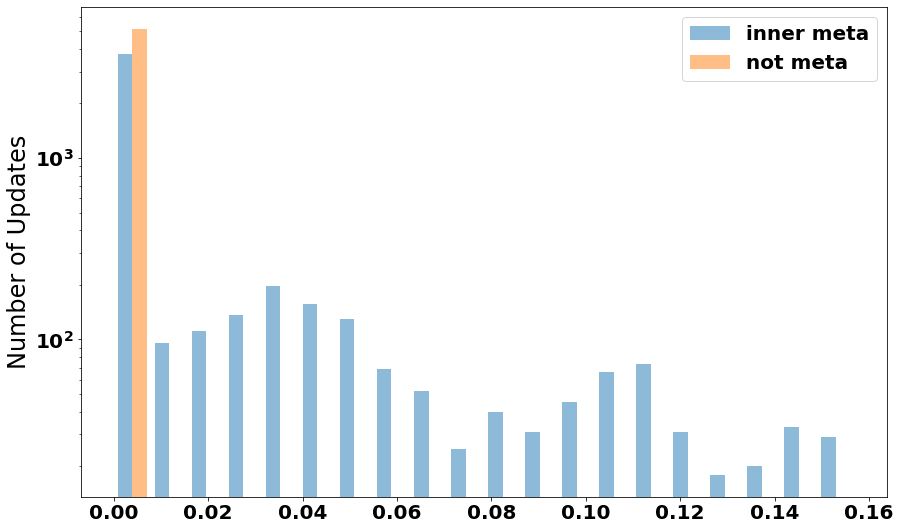}
    \includegraphics[width=.45\linewidth]{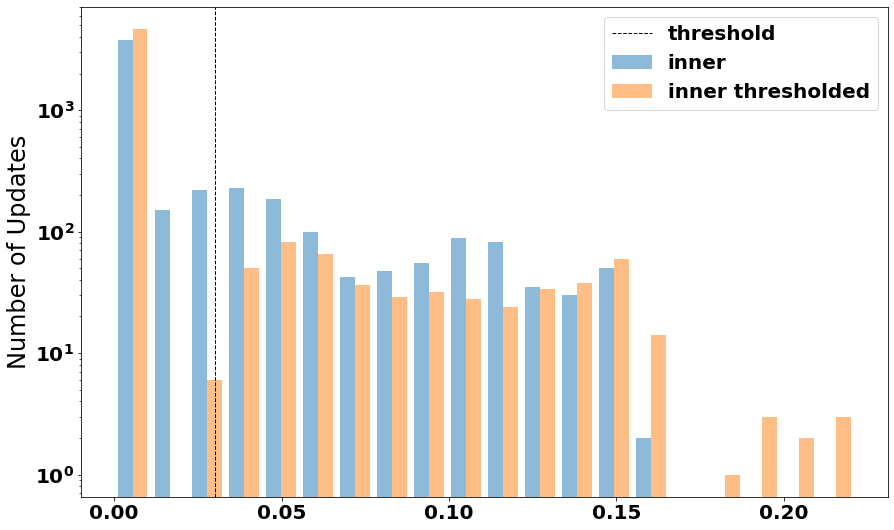}
\caption{A comparison of the weight update magnitudes on Double NMNIST data, shown on a log scale, of an inner loop update and (left) and equivalent not meta trained model update; and (right) an inner loop update that is thresholded. MAML makes fewer non-zero weight updates that are large in magnitude compared to non-meta models. Additionally when thresholded MAML makes fewer non-zero weight updates that are larger in magnitude.}
\label{fig:weight_mag}
\end{figure}
%
% \begin{SCfigure}
%     \includegraphics[width=.4\linewidth]{img/inner_not_meta_log.png}
%     \centering
%     \caption{A comparison of the weight update magnitudes, shown on a log scale, of an inner loop update and an equivalent not meta trained model update. MAML makes fewer non-zero weight updates that are large in magnitude compared to non-meta models.}
%     \label{fig:inner_not_meta}
% \end{SCfigure}
Likewise, we observe that the SNN MAML model only needs a few, large magnitude parameter updates for few-shot learning.
The Table \ref{tab:mag} shows the truncated values of the update magnitude between two training iterations of the output layer for a meta-model and an equivalent non-meta model both trained on Double NMNIST.
The Fig. \ref{fig:weight_mag} gives a more detailed picture by showing histograms of the weight updates.
Comparing the MAML inner loop and the non-meta model's magnitudes, the average update of the inner loop is an order of magnitude larger than the equivalent non-meta model's update.
To summarize, we find that, first, meta-trained models only need one adaptation step to achieve high accuracy when learning a new task (see Table \ref{tab:results}), and second, that these models only need a few updates with a large magnitude to perform few-shot learning (see Table\ref{tab:mag}, Fig. \ref{fig:weight_mag}).
%Given that a meta-trained model needs only one adaptation step to achieve high accuracy when learning a new task as shown in table \ref{tab:results}, table \ref{tab:mag} and Figure \ref{fig:weight_mag} demonstrate that an SNN MAML model only needs few, large magnitude parameter updates for few-shot learning.

Additionally, the magnitude of the weight updates in the inner loop during meta training and adaptation can be thresholded to use even fewer and larger magnitude updates.
During the inner loop adaptation, instead of always updating the parameters the update is gated by a threshold which is described in Eq. (\ref{eq:thresholded}).
\begin{equation}\label{eq:thresholded}
\begin{split}
    \theta_{k}^{n} & =  \begin{cases}
                       \theta_{k}, \mathrm{ if }\Delta w>\Theta,\\
                       \theta_{k-1}, \mathrm{ if }\Delta w<\Theta\\
                      \end{cases}\\
\end{split}
\end{equation}
where $\theta_{k}^{n}$ are the parameters of the model, $\Delta w$ is the magnitude of the update, and $\Theta$ is the threshold.
Thresholding the updates forces the parameters to be larger with fewer updates, which is shown in the Fig. \ref{fig:weight_mag}.
The threshold used in the Fig. \ref{fig:weight_mag}was equal to $5\%$ of the value of the range of magnitude updates.

% \begin{SCfigure}
%     \includegraphics[width=.4\linewidth]{img/inner_thresholded_log.png}
%     \centering
%     \caption{A comparison of the weight update magnitudes, shown on a log scale, of an inner loop update and an inner loop updated that is thresholded. When thresholded fewer non-zero weight updates that are larger in magnitude are made.}
%     \label{fig:inner_thresholded}
% \end{SCfigure}

\subsection{Comparison to Transfer Learning}
\begin{figure}[h!]
\centering
    \includegraphics[width=.45\linewidth]{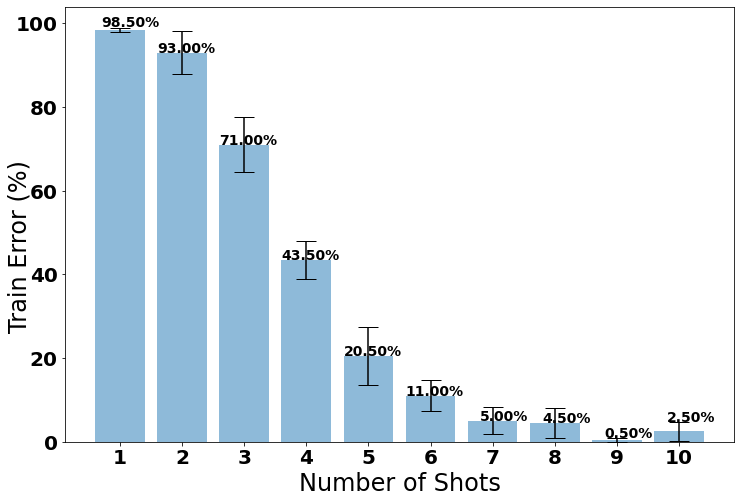}
    \includegraphics[width=.45\linewidth]{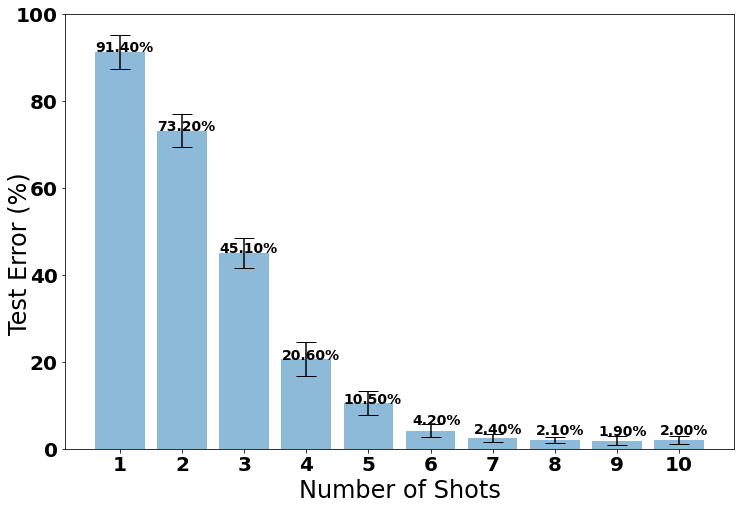}
\caption{The error of a pre-trained non-meta model on Double NMNIST training classes using transfer learning to learn the test set of classes. The model requires more shots than MAML to achieve comparable performance.}
\label{fig:transfer}
\end{figure}
A generalization problem involves learning a function, or model, whose behavior is constrained through a dataset that can make predictions about (\emph{i.e.} learn features than can transfer to) other samples.
A task domain consists of datasets that are related by a common domain, for example, datasets that all consist of double-digit numbers.
Learning on one task in the domain to improve performance on another task is commonly referred to as transfer learning.
Meta-learning can cast transfer learning as a generalization problem because each example, or task, in a given task domain, is a generalization problem instance in meta-learning, which means generalization in meta-learning corresponds to the ability to transfer knowledge between the different problem instances\cite{Andrychowicz_etal16_learto,Vanschoren2019}.
%% Removed because repeated
%Therefore meta-learning casts the problem of transfer learning as a generalization problem that seeks to learn reusable problem structures that are generalizable within a task domain \cite{Andrychowicz_etal16_learto,Vanschoren2019}.

We compare meta-learning to the conventional transfer learning for few-shot learning where a model is pre-trained on a subset of classes within a task domain and then the pre-trained features are transferred to another model that learns to classify new classes within the task domain. 
For comparison to the Double NMNIST SNN MAML model we first pre-trained an SNN model on the 64 classes of the training dataset.
Then we transferred the features to a new model that had an untrained last layer.
The model was trained and tested on 20 of the remaining classes where all layers except the last layer were frozen.
The model was trained on one shot of data at a time and then tested on 20 unseen shots of data.
The few-shot transfer learning results are shown in Fig. \ref{fig:transfer}.
The results shown in the table were averaged over 10 trials.
After the model trains on about 9 or 10 shots of data, the model achieves comparable accuracy to the SNN MAML model shown in Table \ref{tab:results}.
Comparing that to the high accuracy of the SNN MAML models on new tasks after using only one shot of data (see. Fig. \ref{fig:adaptation_steps}, Table \ref{tab:results}.), we can conclude that SNN MAML can adapt to a new task using fewer shots of data. 
%Therefore SNN MAML is able to adapt to new tasks using fewer shots of data than transfer learning because SNN MAML only needs one shot of data and one inner loop adaptation step to achieve high accuracy on new tasks.
%This is because MAML does not need to completely retrain any layers in order to achieve high accuracy on new tasks.

% \begin{table}[h!]
% \begin{center}
% \caption{Number of Shots for a Non MAML Transferred Model to Learn DoubleNMNIST}
% \label{tab:transfer}
% \begin{tabular}{|l|l|l|}
% \hline
% Num Shots & Train Accuracy & Test Accuracy \\ \hline
% 1         & 1.5\pm.5\%            & 8.6\pm3.92\%           \\ \hline
% 2         & 7\pm5.1\%           & 26.8\pm3.8\%          \\ \hline
% 3         & 29\pm6.6\%           & 54.87\pm3.42\%          \\ \hline
% 4         & 56.5\pm4.5\%           & 79.39\pm3.9\%          \\ \hline
% 5         & 79.5\pm6.87\%           & 89.5\pm2.71\%          \\ \hline
% 6         & 89\pm3.74\%           & 95.75\pm1.41\%          \\ \hline
% 7         & 95\pm3.16\%           & 97.57\pm.91\%         \\ \hline
% 8         & 95.5\pm3.5\%          & 97.85\pm.65\%         \\ \hline
% 9         & 99.5\pm.5\%          & 98.08\pm1.06\%         \\ \hline
% 10        & 97.5\pm3.3\%          & 98.03\pm.89\%         \\ \hline
% \end{tabular}
% \end{center}
% \end{table}

\section{Discussion}
Neuromorphic hardware is particularly well suited for online learning at the edge. 
% Scenarios in which online learning can be applied in practice are uncommon, because most learning rules requires a large number of data and iterations.
Here, we demonstrated how to pre-train \acp{SNN} to perform one-shot learning using \ac{MAML}.
The SNN MAML models used the surrogate gradients method to overcome non-linearities that occur in gradient-based training of \acp{SNN}.
We demonstrated our results on combinations of event-based datasets recorded using a neuromorphic vision sensor.

The effective batch size (=1), the precision required for learning from scratch, and the potential correlation in data samples in neuromorphic learning are serious obstacles to deploying neuromorphic learning in practical scenarios.
Fortunately, learning from scratch on the device is generally not even desirable due to robustness and time-to-convergence issues,  especially if the devices are intended for edge applications.
Some form of offline pre-training can alleviate these issues, and \ac{MAML} is an excellent tool to automate this pre-training.

Our results showed that a meta-trained SNN MAML model can learn new event-based tasks in one or a few shots within a task domain.
This enables learning in real-world scenarios when data is streaming, online, and observed only once.
Additionally, the model can relearn prior learned tasks in one or few shots which greatly reduces the impact of catastrophic forgetting because the model does not need to retrain for many iterations.

In hardware for training neural networks, weight updates must be rounded to fall onto values that are resolved at the desired resolution, thereby placing a lower bound to the learning rate.
Conveniently, in few-shot learning, updates are of large magnitude (Fig. \ref{fig:weight_mag}) and a single update must be sufficient to make a change in the output, effectively implying that the learning rate is large.

% One important limitation of \ac{MAML} is its time and space complexity.
% Computing second order gradients as in \ac{MAML} is both memory and compute intensive (second order gradients are one order more expensive than computing gradients). 
% As a result, we were forced to truncate the learning to 30 steps to fit the GPU memory, which limited the temporal credit assignment of our models. 
% Furthermore, the large cost of \ac{MAML} largely prevents learning for a large number of inner loops. 

Meta-training SNN MAML models require considerable computing power and memory. 
The tasks are stored in memory as a sequence of length $T$ with network computations calculated over the sequence.
Our method re-initializes the network dynamics each iteration of training by inputting a partial sequence of a data sample into the network that executes the dynamics but does not update the network.
Additionally, gradients must be computed and stored both in the inner loop and outer loop and backpropagated through the network to meta train. This largely prevents learning for a large number of inner loops. 
While first-order \ac{MAML} requires less memory and compute for each update, it performed significantly worse than \ac{MAML}.
Other first-order MAML methods such as REPTILE \cite{Nichol_etal18_firsmeta} are an alternative that can reduce memory usage at the cost of more compute time and some accuracy.

\subsection{Non-gradient based meta-learning}
MAML is known to learn representations that are general across datasets rather than ``learning to learn'' \cite{Raghu2020Rapid}. 
The results of our experiments with freezing model layers showed this is the case for SNN MAML as well indicating the layers already contain good features at meta-initialization.
Another meta-learning approach done with artificial recurrent neural networks is to train the optimizer itself modeled using an \ac{LSTM} \cite{Andrychowicz_etal16_learto}.
The underlying mechanism relies on the recurrent cell states that capture knowledge that is common across the domain of tasks.

The \ac{SNN} based work \cite{Scherr_etal20_one-lear} falls into this category. There, meta-learning was applied to \acp{SNN} trained using e-prop on arm reach and Omniglot tasks. 
The approach used for the meta-learning combined a Learning Network (LN) to carry out inner loop adaptation and a Learning Signal Generator (LSG) to carry out outer loop generalization that was both modeled by recurrent \acp{SNN}.
% In the arm reach task, the model was trained to reproduce a randomly generated robotic arm movement from one-shot of learning.
% The LSG was shown the target movement to compute a learning signal for the LN, and the LN would perform weight updates to reproduce the target movement.
% During testing, a learning signal was not provided by the LSG.
%In future work we will explore methods of model-based meta-learning that does not directly optimize parameters for specific tasks on a support set and instead conditions the output on a representation of the task to train learning dynamics for rapid learning.

\subsection{Synaptic Plasticity and Meta-Training }
%\textbf{describe superspike/decolle/eprop}

Several recent methods for training \acp{SNN} using gradient descent have been introduced.
% To optimize network parameters, gradient descent incrementally updates the parameters in the direction opposite to the gradient of a loss function. 
%When training \acp{SNN} difficulties arise due to the non-differentiability of the activation function.
% Loss functions for gradient descent are typically defined using the network output at the top layer. 
Assuming a global cost function $\mathcal{L}(s)$ defined on the spikes $s$ of the top layer, the gradients with respect to the weights are:
\begin{equation}\label{eq:loss}
\nabla_{w_{ij}} \mathcal{L} (s^{t}) = 
\frac{\partial \mathcal{L} (s^{t})}{\partial s_{i}^t} 
\frac{\partial s_i^t          }{\partial u_{i}^t} 
\frac{\partial u_i^t          }{\partial w_{ij}} = \frac{\partial \mathcal{L} (s^{t})}{\partial s_{i}^t} \sigma^{'}(u_i^t) p_j^t
\end{equation}
The above equation describes a three-factor learning rule \cite{Gerstner_etal18_eligtrac}.
The factor $\frac{\partial \mathcal{L} (s^{t})}{\partial s_{i}^t}$ describes how changing the output in neuron $i$ modifies the global loss and captures the credit assignment problem.
Interestingly, this learning rule is compatible with synaptic plasticity in the brain so long as there exists a process that computes (or approximates) and communicates $\frac{\partial \mathcal{L} (s^{t})}{\partial s_i^t}$ to neuron $i$. 
Whether and how this can be achieved in a local and biologically plausible fashion is under debate, and there exist several methods to approximate this term on a variety of problems \cite{Zenke_Neftci21_brailear}.
For instance, direct feedback alignment is an important candidate method to overcome this problem in \acp{SNN} \cite{Neftci_etal17_evenranda}. 
\citeauthor{lindsey2020learning} build on direct feedback alignment by meta-training the random parameters of the feedback alignment. 

The relationship of gradient descent with synaptic plasticity means that meta-training is a form of programming of synaptic plasticity.
Programming synaptic plasticity is important for mixed-signal neuromorphic hardware implementation of synaptic plasticity because such circuits are prone to mismatch \cite{Chicca_etal13_neurelec,Prezioso_etal18_spikplas}. 
Even in digital neuromorphic technologies, training programming synaptic plasticity can be useful to approximate an optimal learning rule that would otherwise be impossible or too expensive to implement \cite{Davies_etal18_loihneur}.

\subsection{Meta-learning for Neuromorphic Learning Machines}

% The work by \cite{rosenfeld2021fast} developed an online-within-online meta-learning method via meta-gradient descent that is applicable for learning in neuromorphic edge devices without transfer learning. 
% The method uses a three-factor local learning rule that does not use an offline pre-deployment training with backpropagation.
% This allows a fully online meta learning training method called Online-Within-Online Meta-Learning for SNNs (OWOML-SNN) that performs meta-updates on data streamed to the network.
% To do the meta updates, N SNN models are trained in parallel on N datasets from data sampled by the family of tasks for the outer loop update, and a within-task dataset is used to update an inference SNN in an inner loop update.
% The method follows principles of predictive coding, and does not rely on the backpropagation of gradients.
% They evaluated their method on the Omniglot and MNIST-DVS datasets with 2-way 5 shot learning.
% While \cite{rosenfeld2021fast} avoids pre-training our work shows the potential value of using meta-learning for pre-training models for deployment in neuromorphic hardware enabling higher capacity, accurate fast adaptation to tasks.
% {\color{red} TODO: write how miconi and tianjic are two other ways for Learning and Meta-Training Synaptic Plasticity}

\cite{miconi2018plasticnets} showed that plasticity can be optimized by gradient descent, or meta-learned, in large artificial networks with Hebbian plastic connections called differentiable plasticity.
Network synapses store both a fixed component, a traditional connection weight, and a plastic component as a Hebbian trace that stores a running average of the product of pre and post-synaptic activity modified by a coefficient to control how plastic the synapse is.
Networks using the plastic weights were demonstrated to achieve similar performance to MAML and Matching Networks.
A Hebbian-based hybrid global and local plasticity learning rule similar to the differentiable plasticity presented in \cite{miconi2018plasticnets} was applied with \acp{SNN} in the the Tianjic hybrid neuromorphic chip \cite{wu2020braininspired}.

% Globally plastic network connections learn task specific connections, while the locally plastic network connections learn common features across tasks.
This work meta-optimized Hebbian-based \ac{STDP} learning rules and local meta-parameters on the Omniglot dataset to examine the performance of the model in few-shot learning tasks on the Tianjic neuromorphic hardware.
Our work extends the one \cite{wu2020braininspired} by directly optimizing a putative three-factor learning rule on event-based data and surrogate gradients that could be used for few-shot learning in neuromorphic hardware.

Meta-learning and transfer learning techniques are already presenting themselves as key tools for neuromorphic learning machines.
For example, transfer learning on the Intel Loihi Neuromorphic Research Chip was used to enable few-shot learning \cite{Stewart_etal20_on-cfew-}
There, a gesture classification network was pre-trained using a functional simulator of the Loihi cores. 
The trained parameters were transferred onto the chip. 
Using local synaptic plasticity processors, the hardware was able to learn 5 novel gestures without catastrophic forgetting, achieving 60\% accuracy after a single, 1 second-long presentation of each class.
While encouraging, we believe this performance can be improved using our approach.

\subsection{Conclusion} 
We argued that successful meta-learning on \acp{SNN} holds promise to help reduce training data iterations at the edge, making it essential to designing and deploying neuromorphic learning machines to real-world problems; prevent catastrophic forgetting and learn with low-precision plasticity mechanisms.
As a bi-level learning mechanism, our results point towards a hybrid framework whereby \acp{SNN} are pre-trained  offline for online learning. As a result, we expect a strong redefinition of synaptic plasticity requirements and exciting new learning applications at the edge.

\section*{Data Availability Statement}
The data that supported the findings of this study are found at 
https://github.com/nmi-lab/torchneuromorphic.

\section*{Acknowledgements}
This research was supported by the Intel Corporation (KS, EN), the National Science  Foundation )NSF)  under  grant 1652159 (EN), and the Telluride Neuromorphic Cognition Workshop 2020 (NSF OISE 2020624).
We would like to thank Jan Finkbeiner for his useful comments.

\bibliography{biblio_unique_alt, extra_bib}
\bibliographystyle{jphysicsB}

\end{document}